\documentclass[aos,preprint]{imsart}

\usepackage[OT1]{fontenc}
\usepackage[round]{natbib}
\usepackage[colorlinks,citecolor=blue,urlcolor=blue]{hyperref}
\usepackage{hypernat}

\usepackage{geometry}
\usepackage{amsmath}
\usepackage{amssymb}
\usepackage{color}
\usepackage{tikz}
\usetikzlibrary{bayesnet}
\usepackage{placeins}

\DeclareMathOperator*{\argmax}{arg\,max}
\DeclareMathOperator*{\argmin}{arg\,min}

\newcommand{\secref}[1]{Section~\ref{#1}}
\newcommand{\figref}[1]{Figure~\ref{#1}}
\renewcommand{\eqref}[1]{Equation~\ref{#1}}
\providecommand{\comment}[1]{}

\newtheorem{theorem}{Theorem}
\newtheorem{remark}[theorem]{Remark}

\setattribute{journal}{name}{}

\begin{document}

\begin{frontmatter}
  \title{Detecting Parameter Symmetries in Probabilistic Models}
  \runtitle{Detecting Parameter Symmetries in Probabilistic Models}

  \begin{aug}
    \author{\fnms{Robert} \snm{Nishihara}%
      \ead[label=e1]{rkn@eecs.berkeley.edu}%
      },
    \author{\fnms{Thomas} \snm{Minka}%
      \ead[label=e2]{minka@microsoft.com}%
      }
    \and
    \author{\fnms{Daniel} \snm{Tarlow}%
      \ead[label=e3]{dtarlow@microsoft.com}%
    }%

    \affiliation{UC Berkeley and Microsoft Research}

    \address{Robert Nishihara\\
      University of California\\
      Berkeley, CA 94720, USA\\
      \printead{e1}}
    
    \address{Thomas Minka\\
      Microsoft Research\\
      Cambridge CB1 2FB, UK\\
      \printead{e2}}

    \address{Daniel Tarlow\\
      Microsoft Research\\
      Cambridge CB1 2FB, UK\\
      \printead{e3}}
    
    \runauthor{Nishihara, Minka, and Tarlow}
  \end{aug}

\begin{abstract}
Probabilistic models often have parameters that can be translated,
scaled, permuted, or otherwise transformed without changing the
model. These symmetries can lead to strong correlation and
multimodality in the posterior distribution over the model's
parameters, which can pose challenges both for performing inference
and interpreting the results. In this work, we address the automatic
detection of common problematic model symmetries.  To do so, we
introduce \emph{local symmetries}, which cover many common cases and
are amenable to automatic detection.  We show how to derive algorithms
to detect several broad classes of local symmetries. Our algorithms
are compatible with probabilistic programming constructs such as
arrays, for loops, and if statements, and they scale to models with
many variables.
\end{abstract}
\end{frontmatter}

\section{Introduction}

Probabilistic models play a central role in modern machine
learning. They offer a powerful framework for learning from data, and
they have found applications in a variety of scientific fields beyond
machine learning. A long-standing goal in machine learning and
statistics is to achieve a separation between modeling and inference,
so that users of these tools may focus on specifying models without
having to implement new inference algorithms every time the models
change.  Recently, work in probabilistic programming has taken up this
challenge, seeking to unify probabilistic modeling with computer
programming in order to dramatically increase the effectiveness of
machine learning experts \citep{Darpa2013} and to equip non-experts
with effective tools for specifying models and performing
inference. We anticipate that continued success toward these goals
will decrease the reliance of machine learning practitioners on
tried-and-true models and will shift the community toward a paradigm
grounded in flexible tools for rapidly prototyping and designing new
models \citep{Bishop2013}.

There are many benefits to this increased flexibility. Scientists will be able to swap one model for another without rewriting the inference algorithms.  People will be able to use domain-specific knowledge to build highly-expressive models for their data instead of engineering complex features to pass into a generic algorithm. 

In addition to these benefits, the proliferation of new models
introduces some serious challenges. Foremost perhaps, is the fact that
the available inference algorithms do not work well in all models. One
common cause of failure, which is the focus of this work, is symmetry
(nonidentifiability) in the parameterization of a model. Model
symmetries, which we define formally in \secref{sec:background},
can greatly affect the quality of inference and the interpretability
of the output. We will elaborate on the problems caused by symmetries
in \secref{sec:model_symmetries}. Symmetries have been identified in
common models used in fields from machine learning to econometrics to
political science (see \secref{sec:related_work}), and mechanisms for
breaking these symmetries have been discussed in depth. These
mechanisms depend on the specific type of symmetry under
consideration, but common techniques include reparameterizing the
problem or imposing additional constraints on the parameter values.

If we expect machine learning to be done by experts using
well-understood models, then it is feasible to maintain a catalog of
the symmetries that occur in each common model. If, on the other hand, we
anticipate a wide range of people designing completely original
models, then it is important to automatically detect the presence of
model symmetries so that practitioners can be informed of the problem
or so that the model symmetries can automatically be broken.

To our knowledge, with the exception of some work specific on
permutation symmetries (see \secref{sec:related_work}), this is the
first paper to consider the problem of detecting symmetries in
probabilistic models.

\section{Background} \label{sec:background}
We will represent probabilistic models using factor graphs defined by the pair~$(\boldsymbol\theta, {\bf F})$ of variables~${\boldsymbol\theta} = (\theta_1, \ldots, \theta_N)$ and factors~${\bf F} = (F_1, \ldots, F_K)$.  The factor graph includes all observations, e.g. all data points in a clustering problem.  Some of the variables in~${\boldsymbol\theta}$ represent parameters shared across data points (like the cluster parameters), while others may be per-point hidden variables (like cluster assignments), and others are the observations themselves (which are fixed but still considered part of~${\boldsymbol\theta}$).  We will use~$\Theta$ to denote the space of variable values, so~$\boldsymbol\theta \in \Theta$. The (unnormalized) posterior distribution over the variables given the data can be expressed as the product of the factors
\begin{equation} \label{eq:posterior}
\prod_k F_k({\boldsymbol\theta}) .
\end{equation}
A factor need not depend on all parameters, but we write the product as above for simplicity. Inference tasks typically require us to manipulate the posterior distribution. For instance, we may wish to compute some expectation with respect to the posterior distribution or to compute the posterior marginal distribution of some variable.

A symmetry~$\sigma\colon \Theta \to \Theta$ is a measurable function with a measurable inverse satisfying
\begin{equation} \label{eq:symmetry_definition}
\prod_{k} F_k({\boldsymbol\theta}) \propto  \prod_{k} F_k(\sigma({\boldsymbol\theta}))  ,
\end{equation}
where the product is taken over non-prior terms and the two sides of the equation are viewed as functions of~$\boldsymbol\theta$. We also require that the symmetries fix the components of~$\boldsymbol\theta$ corresponding to observed variables.

From now on, when writing the factors~$F_k$, it will be taken for granted that the factors we consider are not priors. As an aside, the division of factors into likelihood factors and prior factors makes sense in directed models. For arbitrary factor graphs, we can can try to divide the factors into those that get repeated with more data and those that do not repeat. We are interested in symmetries in the factors that repeat, since those are the ones that prevent convergence to a peak.

A statistical model can often be parameterized in different ways and can be expressed with or without hidden variables.  For example, a mixture model can be written with explicit variables that select which cluster a data point came from, or it can be written as a summation over clusters, i.e. with these selectors integrated out.  These different options all lead to different factor graphs, and potentially to different symmetries.  Thus it doesn't make sense to say that a model (in the abstract sense) has symmetries, only that a particular factor graph encoding of that model has symmetries.

\section{The Problem with Parameter Symmetries} \label{sec:model_symmetries}

In Bayesian approaches to machine learning, it is common to make
predictions by manipulating the posterior distribution over the model
parameters~$\boldsymbol\theta$ as expressed in
\eqref{eq:posterior}. In variational approaches to inference, the
exact posterior is approximated by a simpler
function~$q(\boldsymbol\theta)$, which is often factorized and
unimodal. This approximation is effective because the posterior tends
to be increasingly peaked as we receive more and more data. For a
precise statement of this result, see
\citet[Chapter~10.2]{Vandervaart2000}.

Unfortunately, in many real applications, the likelihood function
(which is given by the product of a subset of the factors in the
factor graph) has symmetries in~$\boldsymbol\theta$. For instance, the
likelihood may be invariant to some permutation of the components
of~$\boldsymbol\theta$ (as in mixture models), or the likelihood may
be invariant to adding a constant to certain parameters (as in ranking
models). These symmetries mean that~$\boldsymbol\theta$ is no longer
identifiable from the data, and therefore the posterior does not
become a simple peak as we get more data. Instead, the symmetries of
the likelihood are preserved in the posterior regardless of how much
data we receive.

As a result, variational approximations can be inaccurate, even with
large quantities of data. For example, suppose the likelihood in some
model depends only on~$\theta_1 + \theta_2$. We may observe a lot of
data informing us that~$\theta_1 + \theta_2 = 2$. If the prior
is~$\theta_i \sim \mathcal N(0, 100)$ for~$i \in \{1,2\}$, then the
marginal distribution for either parameter is~$p(\theta_i \,|\,
\mathcal D) = \mathcal N(1, 50)$. If we use a factorized approach that
takes the product of these marginals~$q_1(\theta_1) \, q_2(\theta_2)$
as the approximate posterior and then try to make a prediction
for~$\theta_1 + \theta_2$, we arrive at the very uncertain answer
of~$\mathcal N(2, 100)$. In this example, the problem in the
parameterization is easy to see, but in a more complicated model, the
symmetries can be difficult to find.

Our motivation stems from problems like this one, which is specific to
variational inference. However, there are other reasons to be
concerned about symmetries.

In particular, parameter symmetries impact the reporting of results. A
common practice is to report the mean and variance of each parameter
in isolation. But in the presence of symmetries, these can be
inaccurate or uninformative (in a mixture model, all of the mixture
components may have the same marginals). Visualizing or analyzing the
distance between latent features can be misleading if the different
features are invariant to scaling. For instance, in a collaborative
filtering model based on matrix factorization, any component of the
user features can be scaled as long as the corresponding component of
the item features is scaled in the opposite direction.

More subtle problems can arise from the interpretation of
nonidentifiable parameters. As shown in \citet{Tsiatis1975}, the joint
distribution of the~$K$ potential survival times~$Y_1, \ldots, Y_K$ in
the competing risks model is nonidentifiable from observations
of~$\min_k Y_k$ and~$\argmin_k Y_k$. Overlooking this property may
lead people to mistakenly believe that the $Y_k$'s are independent
when they are not. Symmetries can also lead to confusion when
attempting to compute model evidence, as discussed in
\citet{Neal1999}.

Symmetries do not affect the correctness of sampling algorithms,
though the presence of strong correlation between variables and of
multiple modes can be difficult for many sampling algorithms to
handle. 

\section{Uses for Symmetry Detection} \label{sec:uses_for_symmetry_detection}

As evidenced by the many papers discussing the symmetries present in specific models (see \secref{sec:related_work}), symmetry detection has a variety of uses. Though this paper does not seek to innovate on this front, we nevertheless briefly review several possible uses of automatic symmetry detection. These uses include providing warnings to users of probabilistic programming languages, sanity checking models, and improving inference by automatically breaking model symmetries.

We do not expect all users to be familiar with the problem of model symmetries. By automatically detecting model symmetries, we may help users avoid misinterpreting the results of inference. The warnings that we display can be tailored to the class of model symmetries present as well as to the particular inference algorithm being used. For example, someone running expectation propagation \citep{Minka2001} on a model with permutation symmetries may like to know that the resulting approximate posterior may cover multiple symmetric modes and so its mean and variance may be uninformative. Mean-field approximations may tend to seek individual modes, underestimating posterior variance, and therefore producing reasonable point estimates (even in the presence of symmetries). In such situations, users may try to interpret latent parameters that are in fact nonidentifiable from the data. If the model in question possesses translation or scaling symmetries, we may like to alert the user to the existence of a continuum of comparable point estimates.

When developing a complex model, reporting symmetries may be a useful sanity check. For example, if a parameter meant to denote ``age'' participates in a sign-flip symmetry, something has probably gone wrong in the model specification. More generally, if any variable intended to represent some property of the data is in fact nonidentifiable from the data, there may be a problem.

Automatic symmetry breaking is the most obvious use of automatic
symmetry detection. It is an interesting and important topic, but it
is not the focus of this work. Many approaches have been discussed for
breaking symmetry in specific models with known symmetries. Many of
the techniques used to deal with specific symmetries (such as
constraining or reparameterizing the model) generalize to broad
classes of model symmetries. If used in conjunction, we could
automatically detect symmetries, then automatically break them, then
run inference. Inference algorithms may be better able to summarize
the posterior after the symmetries have been broken. Even sampling
algorithms may perform better as symmetry breaking can reduce the
multimodality and correlation in the posterior. See
\citet{Steyvers2009} for an example of symmetry breaking within Gibbs
sampling. We can also use symmetry breaking as a post-processing
step. Someone doing inference with sampling may wish to project
samples from the symmetric posterior onto a constrained subspace so as
to give more meaningful marginal distributions to the parameters.

There are a number of suggestions from the literature on how to break symmetry. Several examples are discussed in \citet[Chapters~5.3,~5.8,~14.3]{Gelman2007}. Other examples appear in \citet{Bafumi2005, Nobile1998, Stephens2000, Buesing2012, Palomo2007, Erosheva2011, Lopes2004}. But before we can build an automatic symmetry breaker, we need to understand which symmetries we can detect.

\section{Related Work} \label{sec:related_work}

A variety of probabilistic programming languages currently exist or are under development. These languages include BLOG \citep{Milch2007}, BUGS \citep{Lunn2000}, Church \citep{Goodman2008}, Infer.NET \citep{InferNET2012}, and Stan \citep{Stan2013}. Our implementations of the algorithms in this paper operate on models specified in Infer.NET, but the algorithms are general and can be adapted to other model specification languages.

Much effort has been invested in detecting symmetries in constraint-satisfaction problems \citep{Puget2005} and mathematical programs \citep{Liberti2012}. These methods are most closely related to the permutation symmetries that we discuss in \secref{sec:permutation_symmetries} and are typically handled by reducing symmetry detection to a graph automorphism problem. These algorithms can easily be adapted to find permutation symmetries in a probabilistic model.

The lifted inference community is also interested in finding permutation symmetries in models. In this context, symmetries arise from the interplay between a particular model and a particular inference algorithm, and they are viewed as an asset which can be exploited to speed up inference algorithms such as belief propagation \citep{Kersting2009}. In our context, the symmetries that we consider exist independent of how inference is performed. Furthermore, we wish to find and remove symmetries in order to improve the performance of inference algorithms and to improve the interpretability of the results.

One way to avoid parameter symmetries is to directly approximate the predictive distribution, instead of the posterior, as described in \citep{Snelson2005}. Unfortunately, the procedure suggested in their paper is quite expensive. Until a more efficient approach is found, approximating the posterior is still the most practical option.

Symmetries have been identified in specific models arising from
problems in a variety of fields. The Rasch model is used in political
science for ideal point estimation \citep{Rasch1960}, but as it is
commonly implemented, it possesses scaling, sign-flip, and translation
symmetries \citep{Bafumi2005}.  In econometrics, the multinomial
probit model is used to model choice behavior \citep{Geweke1994}. It
too has scaling and translation symmetries
\citep{Nobile1998}. Hierarchical models and mixture models are widely
used in the social sciences \citep{Draper1995} as well as in the
biological sciences \citep{Pritchard2000}. Such models are rife with
symmetries \citep{Gelman2007, Stephens2000}. \citet{Samaniego2010}
considers a material reliability problem in which the stress and
strength parameters exhibit scaling symmetry.

\citet{Neal1998} pointed out that softmax classifiers have a parameter symmetry. He did inference by sampling without breaking the symmetry. Later work adopted the same model in the context of variational inference without breaking the symmetry \citep{Girolami2006, Girolami2007, Kim2006}. \citet{Buesing2012} considered parameter symmetries in linear dynamical systems and proposed a constraint to break them. \citet{Triggs2000} borrowed terminology from physics and described a parameter symmetry as a gauge freedom, and symmetry breaking as gauge fixing. However, they considered optimization problems in which gauge fixing affected only the computational complexity and not the final answer.

\section{Local Parameter Symmetries}

Though we are interested in all symmetries that preserve the
likelihood of the data, we do not imagine that all conceivable
symmetries can be found efficiently. Therefore, we are motivated to
consider classes of symmetries for which efficient detection
algorithms can be constructed. A {\em local symmetry} is a symmetry
that preserves the likelihood at each non-prior factor. In contrast
with \eqref{eq:symmetry_definition}, a local symmetry~$\sigma$
satisfies
\begin{equation} \label{eq:local_symmetry_definition}
 F_k({\boldsymbol\theta}) \propto    F_k(\sigma({\boldsymbol\theta})) ,
\end{equation}
for all non-prior factors~$F_k$. This formulation has two
advantages. First, we believe that it encompasses many of the
symmetries that occur in practice. Second, to compute the local
symmetries of a model, we can consider each factor on its own and
ignore higher-order interactions.

\subsection{Non-local Symmetries}

Not all symmetries are local. Permutation symmetries, which we discuss in \secref{sec:permutation_symmetries}, are quite common and do not fall in this category. Less common and more complex symmetries can arise from mathematical identities. To illustrate this phenomenon, consider the following model. We draw~$v$ from a prior over~$\mathbb R_+$ and~$\theta$ from a prior over~$[0,2\pi)$. We then draw
\begin{eqnarray}
X & \sim & \mathcal N(0, v) \\
Y & \sim & \mathcal N(0, v) 
\end{eqnarray}
independently. Then we observe the data
\begin{eqnarray}
X' & = & X\cos\theta - Y\sin\theta \label{eq:non_local_sym_xprime_eq} \\
Y' & = & X\sin\theta + Y\cos\theta \label{eq:non_local_sym_yprime_eq} . 
\end{eqnarray}
This model possesses the symmetry~$\sigma$ given by
\begin{eqnarray}
\theta & \mapsto & \theta - \alpha \\
X & \mapsto & X\cos\alpha - Y\sin\alpha \\
Y & \mapsto & X\sin\alpha + Y\cos\alpha
\end{eqnarray}
This symmetry preserves the product of the non-prior factors. In particular, it preserves the equalities given in \eqref{eq:non_local_sym_xprime_eq} and \eqref{eq:non_local_sym_yprime_eq}, and it preserves the product
\begin{equation}
\mathcal N(X;0,v) \, \mathcal N(Y;0,v) = \mathcal N(\sigma(X);0,\sigma(v)) \, \mathcal N(\sigma(Y);0,\sigma(v))
\end{equation}
but it does not preserve the individual Gaussian factors.

We believe that symmetries like this one are less common than the local symmetries that we focus on, and we suspect that their presence is more difficult to detect. For these reasons, we do not give an algorithm for detecting more general types of symmetries.

\section{Automatically Detecting Local Symmetries} \label{sec:local_symmetries}

Suppose that we have identified some specific class of transformations of the
model parameters~$T\subset\{f\colon\Theta\to\Theta\}$ (the elements
of~$T$ need not be symmetries) and that we would like to compute the
subset~$ S_T \subset T$ of local symmetries contained in~$T$ defined
by
\begin{equation}
S_T = \{ \sigma \in T \mid \sigma \, \text{is a local symmetry} \} .
\end{equation}
We can compute~$S_T$ as
\begin{equation}
S_T = \bigcap_k \, \{ \sigma \in T \mid  F_k({\boldsymbol\theta}) \propto    F_k(\sigma({\boldsymbol\theta})) \} .
\end{equation}
In order to compute the class of symmetries~$S_T$, we require each
factor to be annotated with the constraints that it imposes on
transformations~$\sigma \in T$. It is important to note that these
annotations need only be made once for each factor by the creators of
the model specification language. There are currently about~$80$
factors in Infer.NET \citep{InferNET2012}, most of which do not
require any annotation (as long as we interpret the lack of any
annotation on a factor to mean that the arguments of that factor may
not be transformed by any symmetry in~$S_T$). In fact, the effort
required to add these annotations is minimal. We will then be able to
compute~$S_T$ by aggregating the constraints from each factor and
solving the resulting system of equations. The ease with which this
formulation gives rise to an algorithm for computing~$S_T$ depends on
the nature of~$T$, but in several useful cases,~$T$ will be a vector
space and the constraints from \eqref{eq:local_symmetry_definition}
will be linear. In such situations,~$S_T$ will also be a vector
space. We now demonstrate this procedure for several specific classes
of symmetries.

\subsection{Scaling Symmetries} \label{sec:scaling_symmetries}

A scaling symmetry is a symmetry~$\sigma$ which multiplies~$\boldsymbol\theta$ pointwise by a vector
\begin{equation}
{\bf v} = (r_1, \ldots, r_N) = (e^{d_1}, \ldots, e^{d_N}) ,
\end{equation}
where each~$r_n \in \mathbb R_+$ and the exponent~$d_n = \log r_n$ (the case of~$r_n \in \mathbb R_-$ is covered by a combination of scaling symmetry and sign-flip symmetry). We will represent this symmetry by the vector of exponents~${\bf d} = (d_1, \ldots, d_N)$. In this case, the underlying space of multiplicative transformations~$T$ is represented by the vector space~$\mathbb R^N$ of exponents.

In order to detect scaling symmetries, we will require each factor to be annotated with the constraints that it imposes on the scaling of its arguments. To illustrate this concept, consider the factor encoding the constraint~$c = a+b$. Any scaling symmetry which hopes to preserve this constraint must scale~$a$,~$b$, and~$c$ by the same amount. In other words, we must have~$d_a = d_b = d_c$. A list of example factors and the constraints they impose is shown in \figref{fig:scaling_constraints}.

\renewcommand{\arraystretch}{2}
\setlength{\tabcolsep}{20pt}
\begin{figure}
\centering
   \begin{tabular}{| c | c | }
     \hline
factor & constraints \\ \hline
$c = a + b$ &$ d_a = d_b = d_c$ \\
$c = a * b$ & $ d_c = d_a + d_b$ \\
$y = \tanh(x)$ & $ d_x = d_y = 0 $ \\
$x \sim \mathcal N(\mu, v)$ & $ d_x = d_{\mu} = \tfrac12 d_v $ \\
$x \ge 0$ & no constraints \\
\hline
   \end{tabular}
\caption{This table shows some example factors and the constraints they impose on potential scaling symmetries.}
\label{fig:scaling_constraints}
\end{figure}

The constraints imposed on~$T$ by some factor are typically linear. Indeed, any integer linear combination of scaling symmetries is again a scaling symmetry. It is possible that a particular factor will impose a nonlinear constraint on~$T$ (for instance, a factor enforcing the constraint that~$\theta_n$ be a power of~$2$), but such factors are rare and are typically not implemented in probabilistic programming languages. Though this framework can accommodate such factors, we will require factors to impose linear constraints on scaling symmetries so that the resulting set of scaling symmetries is a vector space.

Now, if we construct the matrix~${\bf C}$ by stacking together all of the constraints (as row vectors), the scaling symmetries of the model are the vectors in the null space of~${\bf C}$. We can compute the scaling symmetries of the model by constructing the matrix~${\bf C}$ and finding its null space using an algorithm such as Gaussian elimination.

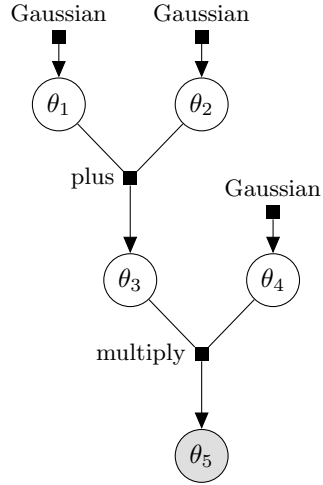
\begin{figure}[ht]
  \begin{center}

\begin{tikzpicture}[x=1.7cm,y=1.8cm]


  \node[obs]       (t5) {$\theta_5$} ; %
  \factor[above=0.5 of t5] {t5-f} {left:multiply} {} {}; %
  \node[latent, above right=0.5 of t5-f]  (t4) {$\theta_4$} ; 
  \node[latent, above left=0.5 of t5-f]  (t3) {$\theta_3$} ; 
  \factor[above=0.5 of t3] {t3-f} {left:plus} {} {}; %
  \node[latent, above right=0.5 of t3-f]  (t2) {$\theta_2$} ; 
  \node[latent, above left=0.5 of t3-f]  (t1) {$\theta_1$} ; 

  \factoredge {t3, t4} {t5-f} {t5} ;
  \factoredge {t1, t2} {t3-f} {t3} ;

  \factor[above=0.25 of t1] {t1-f} {Gaussian} {} {t1} ;
  \factor[above=0.25 of t2] {t2-f} {Gaussian} {} {t2} ;
  \factor[above=0.25 of t4] {t4-f} {Gaussian} {} {t4} ;

\end{tikzpicture}

  \end{center}
  \caption{An example model with scaling symmetries.}
  \label{fig:scaling_example}
\end{figure}

To illustrate the algorithm, we work through an example. Consider the simple model in 
\figref{fig:scaling_example}. Our algorithm ignores the Gaussian priors. It then collects the constraints imposed by each factor on potential scaling symmetries. The ``plus'' factor imposes the constraint~$d_1=d_2=d_3$. The ``multiply'' factor imposes the constraint~$d_3+d_4=d_5$. The observation adds the requirement that~$d_5=0$.

The scaling symmetries of the model are then the solutions to the equation
\renewcommand{\arraystretch}{1}
\setlength{\tabcolsep}{6pt}
\begin{equation}
\bordermatrix{& & & & \cr
                \textcolor{gray}{d_1 = d_3} & 1 &  0  & -1 & 0 & 0 \cr
                \textcolor{gray}{d_2 = d_3} & 0  &  1 & -1 & 0 & 0 \cr
                \textcolor{gray}{d_3 + d_4 = d_5} & 0 & 0 & 1 & 1 & -1 \cr
                \textcolor{gray}{d_5 = 0} & 0 & 0 & 0 & 0 & 1}
\left(\begin{array}{c} d_1 \\ d_2 \\ d_3 \\ d_4 \\ d_5 \end{array} \right) = \left(\begin{array}{c} 0 \\ 0 \\ 0 \\ 0 \end{array} \right) .
\label{eq:scaling_example}
\end{equation}
The rows of the matrix are labeled with the constraints from which they are derived. The solution to 
\eqref{eq:scaling_example} is a one-dimensional vector space given by~$d_1=d_2=d_3=-d_4$, which translates into the space of scaling symmetries given by
\begin{eqnarray}
  \theta_1 & \mapsto & r \theta_1 \\
  \theta_2 & \mapsto & r \theta_2 \\
  \theta_3 & \mapsto & r \theta_3 \\
  \theta_4 & \mapsto & r^{-1} \theta_4 \\
  \theta_5 & \mapsto & \theta_5 .
\end{eqnarray}

\begin{remark}
We mentioned earlier that symmetries are not allowed to change the values of observed variables. However, one of the most common observed values is~$0$, which scaling does not affect. It is often convenient to allow the symmetries we search for to scale observed values that equal~$0$ (doing so allows us to to find the scaling symmetries in the factor~$x\sim\mathcal N(0,v)$ and in the factor~$c = a + 0$ for instance). The same is true for sign-flip symmetries.
\end{remark}

\subsection{Sign-Flip Symmetries} \label{sec:sign_flip_symmetries}

A sign-flip symmetry is a symmetry in which~$\boldsymbol\theta$ is multiplied pointwise by a vector
\begin{equation}
{\bf v}_s=((-1)^{s_1},\ldots,(-1)^{s_N})
\end{equation}
with~$s_n \in \{0,1\}$. As with scaling symmetries, we will represent this symmetry by the binary vector of exponents~${\bf s}=(s_1, \ldots, s_N)$.

Our algorithm for computing sign-flip symmetries is the same as our algorithm for computing scaling symmetries except that some factors impose different constraints and the computations must be done modulo~$2$. As before, each factor must be annotated with the constraints that it imposes on sign-flip transformations. Examples are shown in \figref{fig:sign_flip_constraints}.

\renewcommand{\arraystretch}{2}
\setlength{\tabcolsep}{20pt}
\begin{figure}
\centering
   \begin{tabular}{| c | c | }
     \hline
factor & constraints \\ \hline
$c = a + b$ & $ s_a \equiv s_b \equiv s_c \bmod 2$ \\
$c = a * b$ & $ s_c \equiv s_a + s_b  \bmod 2$ \\
$y = \tanh(x)$ & $ s_x \equiv s_y  \bmod 2$ \\
$x \sim \mathcal N(\mu, v)$ & $ s_x \equiv s_{\mu} \bmod 2$ \,\,\,\, and \,\,\,\, $s_v \equiv 0 \bmod 2$ \\
$x \ge 0$ & $s_x \equiv 0 \bmod 2$ \\
\hline
   \end{tabular}
\caption{This table shows some example factors and the constraints they impose on potential sign-flip symmetries.}
\label{fig:sign_flip_constraints}
\end{figure}

Once again, the sign-flip symmetries of the model form a vector space (over the finite field with~$2$ elements), the constraints imposed by the factors are linear, and the sign-flip symmetries of the model correspond to the null space of the matrix~${\bf C}$ formed by aggregating the constraints from each factor. As before, we can compute the null space of~${\bf C}$ using an algorithm such as Gaussian elimination.

\subsection{Translation Symmetries} \label{sec:translation_symmetries}

We could have defined a translation symmetry as a symmetry~$\sigma$ which translates~$\boldsymbol\theta$ by some vector~${\bf t} = (t_1, \ldots, t_N)$. Then it would follow, as in the case of scaling symmetries and sign-flip symmetries, that each factor must impose linear constraints on potential translation symmetries and that the translation symmetries of the model are the vectors in the null space of the matrix formed by aggregating all of the individual constraints.

Unfortunately, this definition does not capture the full range of symmetries that we feel ought to fall in this category. To illustrate a symmetry that this definition misses, consider the Bayes point machine \citep{Girolami2006}. This model maintains latent vectors~${\bf w}_1, \ldots, {\bf w}_K$ for each of~$K$ classes, and it labels a data point~${\bf x}$ with label~$y$ by computing a noisy score
\begin{equation}
s_k = {\bf w}_k^{\mathsf T}{\bf x} + \epsilon_k ,
\end{equation}
for each class (where~$\epsilon_k \sim \mathcal N(0, v)$ represents the noise), and setting
\begin{equation}
y = \argmax_k s_k .
\end{equation}

Given any vector~${\bf v}$ of the appropriate dimension, the translation
\begin{equation}
({\bf w}_1, \ldots, {\bf w}_K) \mapsto ({\bf w}_1 + {\bf v}, \ldots, {\bf w}_K + {\bf v})
\end{equation}
does not change the likelihood of the data. So far, this translation appears to satisfy our preliminary definition of a translation symmetry. However, the vectors~${\bf w}_k$ are not the only parameters in the model. For the purpose of running inference, the model that we construct will also maintain latent variables for the scores~$s_k$. When we translate each~${\bf w}_k$ by~${\bf v}$, we must translate each~$s_k$ by~${\bf v}^{\mathsf T}{\bf x}$. If we wish to categorize such symmetries as translation symmetries, we must allow the vector~${\bf t}$ by which we translate~$\boldsymbol\theta$ to depend on~$\boldsymbol\theta$. Because of this dependence, translation symmetries differ from scaling and sign-flip symmetries.

With this example in mind, we define a translation symmetry to be a symmetry~$\sigma$ satisfying
\begin{equation}
\sigma(\boldsymbol\theta) = \boldsymbol\theta + {\bf t}(\boldsymbol\theta) ,
\end{equation}
where~${\bf t}(\boldsymbol\theta) = (t_1(\boldsymbol\theta), \ldots, t_N(\boldsymbol\theta))$ such that~${\bf t}(\boldsymbol\theta)$ can only depend on the parameters not being translated (i.e.~on the~$\theta_n$ such that~$t_n(\boldsymbol\theta) = 0$). To motivate this requirement, we note that it leads to the result that the existence of a translation symmetry~$\sigma$ as above implies the existence of a family of translation symmetries
\begin{equation}
\sigma_c(\boldsymbol\theta) = \boldsymbol\theta + c{\bf t}(\boldsymbol\theta)
\end{equation}
satisfying properties such as
\begin{equation}
\sigma_{c_1+c_2}(\boldsymbol\theta) = \sigma_{c_2}(\sigma_{c_1}(\boldsymbol\theta)) ,
\end{equation}
at least for integer~$c,c_1,c_2$. The existence of this family is consistent with an intuitive notion of the properties that a translation symmetry ought to satisfy.

As before, each factor must be annotated with the constraints that it imposes on potential translation symmetries. Examples are shown in \figref{fig:translation_constraints}.

\renewcommand{\arraystretch}{2}
\setlength{\tabcolsep}{20pt}
\begin{figure}
\centering
   \begin{tabular}{| c | c | }
     \hline
factor & constraints \\ \hline
$c = a + b$ & $t_c = t_a + t_b$ \\
$c = a * b$ & $t_c = t_a b + t_b a + t_at_b$ \,\,\,\, and \,\,\,\, $t_at_b = 0$ \\
$y = \tanh(x)$ & $t_x = t_y = 0$ \\
$x \sim \mathcal N(\mu, v)$ & $t_x = t_{\mu}$ \,\,\,\,  and \,\,\,\,  $t_v = 0$ \\
$x \ge 0$ & $t_x = 0$ \\
\hline
   \end{tabular}
\caption{This table shows some example factors and the constraints they impose on potential translation symmetries.}
\label{fig:translation_constraints}
\end{figure}
\renewcommand{\arraystretch}{1}

Therefore, we can compute the translation symmetries of the model by aggregating all of the constraints and solving the resulting (nonlinear) system of equations. The translation symmetries of a model can be written as a union of vector spaces, and we can compute this set by solving several separate linear systems. Since we allow the constraints to depend on~$\boldsymbol\theta$, the coefficients in the resulting system of equations may be symbolic.

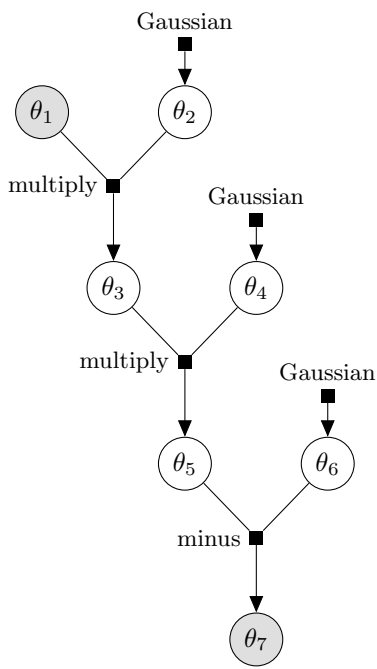
\begin{figure}[ht]
  \begin{center}

\begin{tikzpicture}[x=1.7cm,y=1.8cm]


  \node[obs]                     (t7)  {$\theta_7$} ; %
  \factor[above=0.5 of t7] {t7-f} {left:minus} {} {} ;
  \node[latent, above right=0.5 of t7-f]       (t6) {$\theta_6$} ; %
  \node[latent, above left=0.5 of t7-f]       (t5) {$\theta_5$} ; %
  \factor[above=0.5 of t5] {t5-f} {left:multiply} {} {}; %
  \node[latent, above right=0.5 of t5-f]  (t4) {$\theta_4$} ; 
  \node[latent, above left=0.5 of t5-f]  (t3) {$\theta_3$} ; 
  \factor[above=0.5 of t3] {t3-f} {left:multiply} {} {}; %
  \node[latent, above right=0.5 of t3-f]  (t2) {$\theta_2$} ; 
  \node[obs, above left=0.5 of t3-f]  (t1) {$\theta_1$} ; 

  \factoredge {t5, t6} {t7-f} {t7} ;
  \factoredge {t3, t4} {t5-f} {t5} ;
  \factoredge {t1, t2} {t3-f} {t3} ;

  \factor[above=0.25 of t2] {t2-f} {Gaussian} {} {t2} ;
  \factor[above=0.25 of t4] {t4-f} {Gaussian} {} {t4} ;
  \factor[above=0.25 of t6] {t6-f} {Gaussian} {} {t6} ;

\end{tikzpicture}

  \end{center}
  \caption{An example model with translation symmetries.}
  \label{fig:translation_example}
\end{figure}

To illustrate the way in which translation symmetries may arise, consider the simple model shown in \figref{fig:translation_example}. In the model,~$\theta_2$,~$\theta_4$, and~$\theta_6$ are drawn from Gaussian priors, and~$\theta_1$ and~$\theta_7$ are observed. Our algorithm ignores the Gaussian priors and finds the translation symmetries
\begin{eqnarray}
  \theta_1 & \mapsto & \theta_1\\
  \theta_2 & \mapsto & \theta_2\\
  \theta_3 & \mapsto & \theta_3 \\
  \theta_4 & \mapsto & \theta_4 + t\\
  \theta_5 & \mapsto & \theta_5 + \theta_3 t\\
  \theta_6 & \mapsto & \theta_6 + \theta_3 t \\
  \theta_7 & \mapsto & \theta_7 
\end{eqnarray}
for all $t \in \mathbb R$, and
\begin{eqnarray}
  \theta_1 & \mapsto & \theta_1\\
  \theta_2 & \mapsto & \theta_2 + t\\
  \theta_3 & \mapsto & \theta_3 + \theta_1 t\\
  \theta_4 & \mapsto & \theta_4 \\
  \theta_5 & \mapsto & \theta_5 + \theta_1 \theta_4 t\\
  \theta_6 & \mapsto & \theta_6 + \theta_1 \theta_4 t \\
  \theta_7 & \mapsto & \theta_7 
\end{eqnarray}
for all $t \in \mathbb R$.

\section{Permutation Symmetries} \label{sec:permutation_symmetries}

A permutation symmetry is a symmetry~$\sigma$ that permutes the
components of~$\boldsymbol\theta$. Such symmetries arise in models
with multiple interchangeable units such as the mixture components in
a mixture of Gaussians, the topics in Latent Dirichlet Allocation
\citep{Blei2003}, the hidden units in a neural network, and the latent
features in a collaborative filtering model based on matrix
factorization.

In a mixture model with symmetric priors, all mixture components have
the same marginal distributions. This property can make it difficult
to summarize the output of a sampler. This is known as the
label-switching problem. Papers on Latent Dirichlet Allocation
sometimes avoid this by using only one sample
\citep{Griffiths2004}. Whether or not permutation symmetry poses a
problem for variational inference can depend on how well separated the
mixture components are. If the mixture components are well-separated,
variational inference tends to get stuck in one mode and so isn't
bothered by the symmetry. The posterior distribution is inaccurate,
but not in a way that significantly impacts test results.

We give an algorithm for detecting permutation symmetries by reduction
to a graph automorphism computation, a problem for which there is no
known polynomial time algorithm. This approach is very similar to
existing work in the constraint satisfaction and mathematical
programming literature \citep{Puget2005, Liberti2012}, and we do not
consider this section to be novel. In
\secref{sec:for_loops_with_permutation_symmetries}, we will give an
efficient algorithm for finding many of the permutation symmetries
that arise in practice by making use of arrays.

\subsection{Detecting Permutation Symmetries}

More precisely, we will label the edges and vertices of the factor
graph so that the maps from the factor graph to itself that preserve
the edge and vertex labels correspond to the permutation symmetries of
the model that can be deduced without any knowledge of the
mathematical properties of the factors other than symmetry of each
factor with respect to its arguments (for instance, the factor
encoding the constraint~$c=a+b$ is symmetric with respect to~$a$
and~$b$, but not with respect to~$a$ and~$c$).

Each type of factor must be annotated with its identity (for instance, all binary plus factors have the same label, but a binary plus factor and a ternary plus factor have different labels). Each factor must also associate a label with each of its arguments such that two arguments have the same label if and only if the factor is symmetric with respect to those two arguments (we will refer to the information contained in a factor's argument labels as the ``symmetry structure'' of that factor). Now, in the factor graph, label each factor with its identity, and for each edge connecting variable~$\theta_n$ to a factor~$F_k$, label the edge with the label that factor~$F_k$ associates with the argument~$\theta_n$.

The permutation symmetries that we can deduce from these annotations
(factor labels and factor symmetry structures) are the permutation
symmetries~$\sigma$ such that
\begin{equation}
\prod_{\text{label}(k) = c} f_k(\sigma(\boldsymbol\theta)) = \prod_{\text{label}(k) = c} f_k(\boldsymbol\theta) ,
\end{equation}
for all factor labels~$c$.  These permutation symmetries correspond to
the automorphisms of the factor graph that preserve the factor and
edge labels.

Now, we can take our factor graph and compute its automorphism group
\citep{Mckay1981,Darga2004}. As is common with these computations, the
group of permutation symmetries that we compute can be compactly
represented by a set of generators (symmetries that can be composed to
produce the entire set). Despite the fact that the problem of
detecting the existence of nontrivial graph automorphisms is not known
to be solvable in polynomial time \citep{Kobler1993}, we expect it to
perform well in practice because graph automorphism can be tested in
linear time for almost all graphs \citep{Babai1980}. However, as we
describe in \secref{sec:for_loops_with_permutation_symmetries}, we
will ultimately compress the factor graphs using arrays (plate
notation) so that the graphs whose automorphisms we attempt to compute
are very small (possessing on the order of several dozen vertices).

Note that, with this approach, it is important to write models so that
the symmetries are not hidden because this approach will not examine
the interactions between factors.  This approach will find the
symmetry of~$a + b + c$, but it will not find the symmetry of~$(a + b)
+ c$, where the first refers to an addition factor with three
arguments and the second refers to the sequential use of two addition
factors. Therefore, when summing more than two variables, it will be
important to use an~$n$-ary version of the sum factor (we point out
that it is not possible to find all of the symmetries of~$(a + b) + c$
using only the information that we considered, namely the
commutativity of addition, in this case, associativity is necessary as
well, which we did not take into account). This problem does not arise
with the other forms of symmetry that we have discussed in this paper.

\section{Arrays, For Loops, and If Statements} \label{sec:for_loops_if_statements}

Probabilistic programs often include support for conventional programming constructs such as arrays, for loops, and if statements, so it is important for our algorithms to be compatible with models specified using these objects.

\subsection{Arrays} \label{sec:for_loops}

The detection methods, as we described them so far, operate on the full factor graph. A matrix factorization problem with millions of users, thousands of items, and hundreds of underlying features will have an unmanageably large factor graph. However, such large models do not have billions of different variables, each with their own unique interpretations. These models typically have billions of variables many of which share similar modeling roles (indeed, specifying the model in code requires on the order of one hundred lines as opposed to one billion). When defining a model in code, it is generally convenient to declare all of these similar variables at once using an array (this is analogous to drawing a graphical model using plate notation).

Intuitively, the symmetries of a model should not depend on the sizes of the arrays involved (though it is possible to contrive examples of symmetries that appear only for arrays of certain sizes). For instance, the translation symmetries in the Bayes point machine persist when we add or remove data and when we add or remove classes. For this reason, we ought to be able to formulate an algorithm for detecting the symmetries of our model with complexity independent of the sizes of the arrays involved. If we can adapt our algorithms to enable us to treat arrays of variables as individual variables, then we will have achieved this goal.

Note that the performance enhancement obtained from treating arrays as individual variables arises from the presumed repetitive structure that arrays represent. This assumption holds whenever arrays are used to describe the repetition in a graphical model arising from plate notation. However, a programmer could choose to create a model with many variables (each with its own distinct role) and store them all in an array. In such a situation, the presumed structure does not exist, and there is no computational benefit to be gained from the array. In order to be precise about the guarantees of our array-boosted algorithms, we must either allow our algorithms to default to operating on the full factor graph or we must consider the possibility that we will not detect all symmetries. Both options have their advantages, and it may be the case that some combination of the two is ideal. However, due to the potential size of the full factor graph, we choose to devise an algorithm that never unrolls the factor graph and that instead bounds the set of symmetries between a subset and a superset.

We must adapt our definitions of the various classes of symmetries at the coarser resolution of arrays. We do this by treating arrays as individual variables. In the case of translation symmetries, the restriction we place on~$t({\boldsymbol\theta})$ applies at the level of arrays (if a variable in an array is translated, then no variable in that array can affect the extent to which another variable is translated). In the case of permutation symmetries, we allow the permutation of entire arrays and the permutation of the indices within an array, but we do not allow permutations that would send some variables in the same array to different arrays.

\subsubsection{Arrays with Scaling, Sign-Flip, and Translation Symmetries}

Let~$(D_1, \ldots, D_N)$ be a representation of a scaling, sign-flip, or translation symmetry of the model. For concreteness, we will imagine that the vector consists of the exponents representing a scaling symmetry. If the variable~$\theta_n$ is an array, then~$D_n$ will be an array of exponents. Suppose that the arrays in the model are collectively indexed by the indices~$i_1, \ldots, i_R$, where~$i_r \in I_R$ for each~$r$. Then we will use~$D_n[I]$ to indicate the element of~$D_n$ indexed by~$I \in I_1 \times \cdots \times I_R$. If~$D_n$ is not indexed by some (or any) of the~$i_r$, then we simply ignore those indices. We detect symmetries in this context as follows.
\\
\\
\noindent{\bf The Simple Case:} Suppose that for some model and some class of symmetries, the set of constraints imposed by the factors can be partitioned into~$J$ subsets corresponding to the functions~$f_1, \ldots, f_J$, where the set of constraints corresponding to~$f_j$ is given by
\begin{equation}
f_j\left(D_1[I], \ldots, D_N[I]\right) = 0 \label{eq:array_one_condition} 
\end{equation}
for all~$I$. We proceed by solving the system of equations given by
\begin{equation}
V = \{ (v_1, \ldots, v_N) \mid f_j(v_1, \ldots, v_N) = 0 \,\, \text{for all} \, j \} \label{array_one_condition_variables}.
\end{equation}
Importantly, \eqref{array_one_condition_variables} treats arrays of variables as individual variables and can be solved without unrolling the full factor graph. Now we can compactly represent the space of symmetries as
\begin{equation}
\{ (D_1, \ldots, D_N) \mid (D_1[I], \ldots, D_N[I]) \in V \,\, \text{for all} \, I  \} .
\end{equation}
\\
\noindent{\bf Partial Solution to the General Case:} Now suppose that we cannot partition the constraints into subsets of the form described in \eqref{eq:array_one_condition}. For instance, some constraint may single out some specific index in one of the~$D_n$'s, or some constraint may relate multiple elements of one of the~$D_n$'s. Let~$D$ be the subvector of ~$(D_1, \ldots, D_N)$ consisting of the arrays~$D_n$ for which some constraint singles out an individual element or relates multiple elements. Requiring~$D_n$ to be constant for each~$D_n \in D$ (in the sense that~$D_n[I]$ does not change as we vary~$I$) allows us to write the constraints as in \eqref{eq:array_one_condition} and places us back in the simple case. Solving for the space~$V$ as before, we can find and represent all symmetries~$(D_1, \ldots, D_N)$ subject to the constraint that~$D_n$ is constant for each~$D_n \in D$.

This algorithm imposes additional constraints and therefore finds a
subset of the symmetries. We can find a superset of the symmetries by
following the same procedure but ignoring the constraints that relate
multiple elements of the same array (or index specific
elements). Though we have not implemented this suggestion, we
speculate that we can then tighten the superset by attempting to prune
the resulting solutions that are inconsistent with the ignored
constraints (it may make sense to invest effort in doing this for
problematic factors that occur frequently).

\subsubsection{Arrays with Permutation Symmetries} \label{sec:for_loops_with_permutation_symmetries}

Permutation symmetries interchange variables with similar roles, and variables with similar roles are often declared using arrays. Arrays in Infer.NET \citep{InferNET2012} are indexed by ``range'' objects (for instance, in a neural network example, one range may index the data set, one range may index the components of a data point, and one range may index the hidden units). In many common mixture model examples, permutation symmetries correspond to the permutation of the indices of a single range (the hidden units in this case).

A given range may index multiple arrays. For instance, the range corresponding to hidden units must index the array storing the weights into the hidden units, the array storing the weights out of the hidden units, and the array storing the biases of the hidden units (at least). In general, the indices of a range will be permutable if each factor is symmetric with respect to permutations of the indices of that range (this is often the case, although it would not be the case for some factors like a cumulative sum factor or a factor that accesses a specific element of the array) and if the range does not index an array containing observed values. If an array is indexed by a random variable, we check to see if any factor prevents the permutation of the values taken on by the random index (and if so, the permutation of the corresponding range is disallowed).

Now, we restrict our search for permutation symmetries to two separate searches, one for permutations of variables as before (but now treating arrays of variables as individual variables) and one for permutations of the indices of a range.

Note that when we treat an array of variables as a single variable and then proceed to search for permutation symmetries, we exclude the possibility of finding permutation symmetries that send elements of the same array to different arrays (something that makes little semantic modeling sense). In return, the complexity of our permutation detection algorithm becomes independent of the sizes of the arrays. We believe that this trade-off is a reasonable one.

\subsection{If Statements} \label{sec:if_statements}

Gate notation can be used to describe context-sensitive independence in factor graphs \citep{Minka2008} and are one way of implementing if statements and switch statements in probabilistic programs.  All of the symmetry detection algorithms that we have described are compatible with gates. Consider the example
\begin{verbatim}
    bool b = Bernoulli(0.5)
    if (b)
        x = y
    else
        x = z
\end{verbatim}
Local symmetries must be compatible with the constraints imposed by the factors in both the if block and the else block. In this example, in the case of scaling symmetries, the if block imposes the constraint~$d_x = d_y$, and the else block imposes the constraint~$d_x=d_z$. The if-else block could equivalently be rewritten as a pair of factors, and by doing so, we would arrive at the same constraints.

This procedure will produce no false positives, but there are situations in which it will be overly restrictive. For instance, if the condition of some if block is always false, there is no need to impose the constraints from the factors in that block. If an if block and the corresponding else block both use a temporary variable with the same name, and if this variable is never accessed outside of the if-else block, then the two instances of the variable ought to be treated as separate variables. Consider the following example.
\begin{verbatim}
    bool b = Bernoulli(0.5)
    double x
    if (b)
        x = Gaussian(0, 1)
        y1 = x + 1
    else
        x = Gaussian(0, 1)
        y2 = x ^ 2
\end{verbatim}
The if block prevents any symmetry from scaling \texttt{x}, and the else block prevents any symmetry from translating \texttt{x}. As a consequence, no symmetries will be able to scale or translate \texttt{y1} or \texttt{y2}, the variable of interest. But if \texttt{x} is a temporary variable that is never used outside of the if-else block, then the two instances of \texttt{x} ought to be treated as different variables. To solve this problem, we could rewrite the code as below.
\begin{verbatim}
    bool b = Bernoulli(0.5)
    double x1, x2
    if (b)
        x1 = Gaussian(0, 1)
        y1 = x1 + 1
    else
        x2 = Gaussian(0, 1)
        y2 = x2 ^ 2
\end{verbatim}
Doing so would enable us to find symmetries that translate \texttt{y1} and scale \texttt{y2}.

In a probabilistic programming language that allows users to create very general programs, these kinds of edge cases are bound to arise. Such edge cases will not lead our algorithms to produce any false positives, but they can lead to the omission of some symmetries. This is an argument for running symmetry detection algorithms after performing various program canonicalizations (such as pruning unreachable code and renaming variables). In fact, 
these program transformations are often already done within inference engines for other reasons.

\section{Some Demonstrations} \label{sec:demonstrations}

We have evaluated our algorithms on a number of models, including a neural network, Latent Dirichlet Allocation, and a mixture of Gaussians model. Our algorithms discovered all of the symmetries in these models. In this section, we give a detailed description of three more models and the symmetries that our algorithms detect in these models. These examples showcase the capacity of our algorithms to find symmetries in complicated models and also highlight some of the limitations of our algorithms.

\subsection{Multinomial probit classifier}

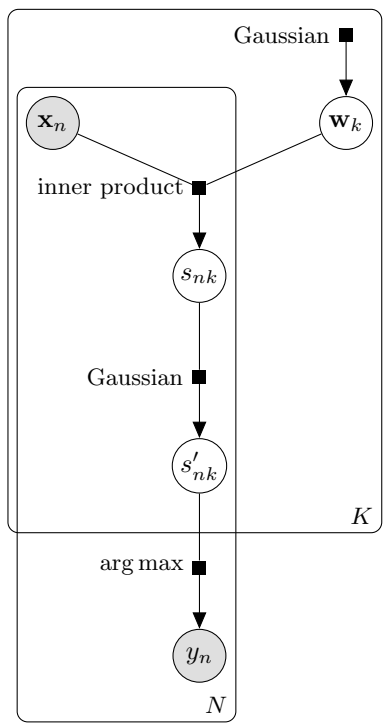
\begin{figure}[t]
  \begin{center}

\begin{tikzpicture}[x=1.7cm,y=1.8cm]


  \node[obs]                      (Y)      {$y_n$} ; %
  \node[latent, above=of Y]       (S')     {$s_{nk}'$} ; %
  \node[latent, above=of S']      (S)      {$s_{nk}$} ; %
  \node[latent, above right=1.2 of S]       (W)      {${\bf w}_k$} ; %
  \node[obs, above left=1.2 of S]          (X)      {${\bf x}_n$} ; %

  \factor[above=of W] {w-f} {left:Gaussian} {} {W} ; %
  \factor[above=of S] {s-f} {left:inner product} {X,W} {S} ; %
  \factor[above=of S'] {s'-f} {left:Gaussian} {S} {S'} ; %
  \factor[above=of Y] {y-f} {left:$\argmax$} {S'} {Y} ; %

  \plate {x} { %
    (X) %
    (S)(s-f)(s-f-caption) %
    (S')(s'-f)(s'-f-caption) %
    (Y)(y-f)(y-f-caption) %
  } {$N$} ;
  \plate {w} { %
    (x.north west)
    (W)(w-f)(w-f-caption) %
    (S)(s-f)(s-f-caption) %
    (S')(s'-f)(s'-f-caption) %
  } {$K$} ;

\end{tikzpicture}

  \end{center}
  \caption{Factor graph for the multinomial probit classifier.}
  \label{fig:bayes_point_machine_factor_graph}
\end{figure}

The multinomial probit classifier, which is described in \citet{Girolami2006}, maintains latent vectors~${\bf w}_1, \ldots, {\bf w}_K$ for each of~$K$ classes. It assigns a data point~${\bf x}_n$ to class~$y_n$ by computing a noisy score via
\begin{eqnarray}
s_{nk} & = & {\bf w}_k^{\mathsf T}{\bf x}_n \\
s_{nk}' & \sim & \mathcal N(s_{nk}, 1)
\end{eqnarray}
for each class, and setting
\begin{equation}
y_n = \argmax_k s_{nk}' .
\end{equation}
The factor graph for this model is shown in \figref{fig:bayes_point_machine_factor_graph}.

This model has no permutation, scaling, or sign-flip symmetries. However, our algorithm finds the translation symmetries given by
\begin{eqnarray}
{\bf w}_k & \mapsto & {\bf w}_k + {\bf v} \\
s_{nk} & \mapsto & s_{nk} + {\bf v}^{\mathsf T} {\bf x}_n \\
s_{nk}' & \mapsto & s_{nk}' + {\bf v}^{\mathsf T} {\bf x}_n 
\end{eqnarray}
for all vectors~${\bf v}$ of the same dimension as the data.

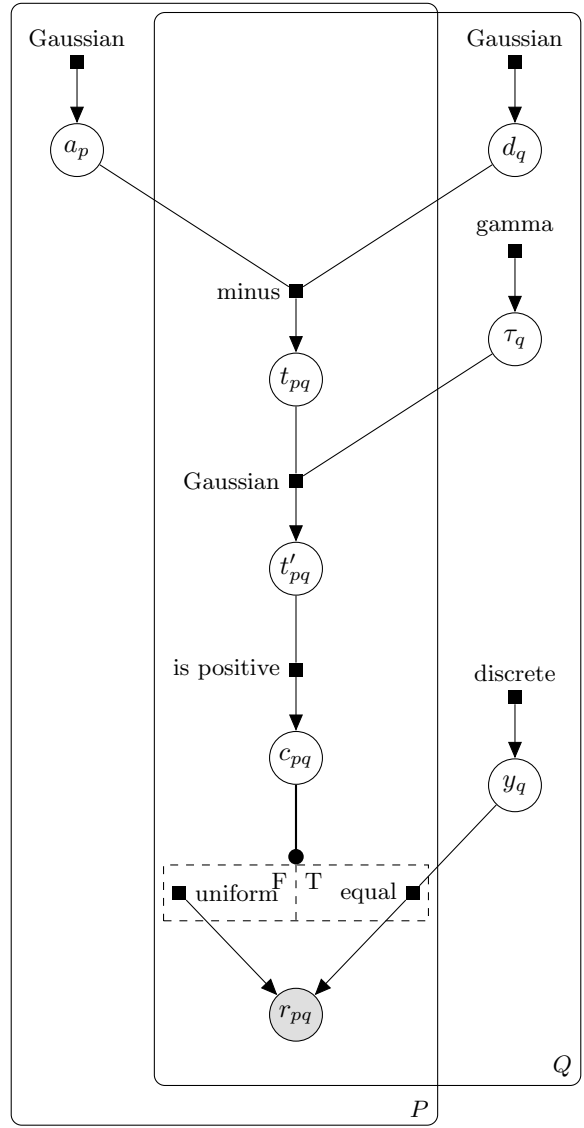
\begin{figure}[!ht]
  \begin{center}

\begin{tikzpicture}[x=1.7cm,y=1.8cm]


  \node[obs]                            (response)  {$r_{pq}$} ; %
  \node[latent, above right=2.0 of response]       (trueAnswer)  {$y_{q}$} ; %
  \node[latent, above=1.5 of response]      (correct)  {$c_{pq}$} ; %
  \node[latent, above=of correct]       (noisyAdvantage) {$t_{pq}'$} ; %
  \node[latent, above right=2.0 of noisyAdvantage]       (discrimination)      {$\tau_q$} ; %
  \node[latent, above=of noisyAdvantage]       (advantage)      {$t_{pq}$} ; %
  \node[latent, above left=2.0 of advantage]       (ability)      {$a_{p}$} ; %
  \node[latent, above right=2.0 of advantage]       (difficulty)      {$d_{q}$} ; %

  \factor[above=of ability] {ability-f} {Gaussian} {} {ability} ; %
  \factor[above=of difficulty] {difficulty-f} {Gaussian} {} {difficulty} ; %
  \factor[above=of discrimination] {discrimination-f} {gamma} {} {discrimination} ; %
  \factor[above=of advantage] {advantage-f} {left:minus} {ability,difficulty} {advantage} ; %
  \factor[above=of noisyAdvantage] {noisyAdvantage-f} {left:Gaussian} {advantage,discrimination} {noisyAdvantage} ; %
  \factor[above=of correct] {correct-f} {left:is positive} {noisyAdvantage} {correct} ; %
  \factor[above right=1 of response] {equal-f} {left:equal} {trueAnswer} {response} ; %
  \factor[above left=1 of response] {uniform-f} {right:uniform} {} {response} ; %
  \factor[above=of trueAnswer] {trueAnswer-f} {discrete} {} {trueAnswer} ; %

  \vgate {response-vgate} %
  {(equal-f)(equal-f-caption)} {F} 
  {(uniform-f)(uniform-f-caption)} {T} 
  {correct} ; %

  \plate {question} { %
    (response-vgate)
    (difficulty)(difficulty-f)(difficulty-f-caption) %
    (discrimination)(discrimination-f)(discrimination-f-caption) %
    (discrimination) %
    (advantage)(advantage-f)(advantage-f-caption) %
    (noisyAdvantage)(noisyAdvantage-f)(noisyAdvantage-f-caption) %
    (correct)(correct-f)(correct-f-caption) %
    (trueAnswer)(trueAnswer-f)(trueAnswer-f-caption) %
    (uniform-f)(uniform-f-caption)
    (equal-f)(equal-f-caption)
    (response) %
  } {$Q$} ;
  \plate {student} { %
    (response-vgate)
    (question.south)
    (question.north)
    (ability)(ability-f)(ability-f-caption) %
    (advantage)(advantage-f)(advantage-f-caption) %
    (noisyAdvantage)(noisyAdvantage-f)(noisyAdvantage-f-caption) %
    (correct)(correct-f)(correct-f-caption) %
    (uniform-f)(uniform-f-caption)
    (equal-f)(equal-f-caption)
    (response) %
  } {$P$} ;

\end{tikzpicture}

  \end{center}
  \caption{Factor graph for difficulty versus ability.}
  \label{fig:difficulty_ability_factor_graph}
\end{figure}

As pointed out in section~\ref{sec:background}, the way that a model is expressed as a factor graph can affect its symmetries.  We illustrate this by writing the multinomial probit model differently, using an additional set of hidden variables representing the additive noise
\begin{eqnarray}
s_{nk} & = & {\bf w}_k^{\mathsf T}{\bf x}_n \\
u_{nk} & \sim & \mathcal N(0, 1) \label{eq:u_prior} \\
s_{nk}' &=& s_{nk} + u_{nk}
\end{eqnarray}
If we consider \eqref{eq:u_prior} to be a prior factor, then we now have a scaling symmetry
\begin{eqnarray}
{\bf w}_k & \mapsto & r {\bf w}_k \\
s_{nk} & \mapsto & r s_{nk} \\
u_{nk} & \mapsto & r u_{nk} \\
s_{nk}' & \mapsto & r s_{nk}'
\end{eqnarray}
and the larger set of translation symmetries
\begin{eqnarray}
{\bf w}_k & \mapsto & {\bf w}_k + {\bf v}_k \\
s_{nk} & \mapsto & s_{nk} + {\bf v}_k^{\mathsf T}{\bf x}_n \\
u_{nk} & \mapsto & u_{nk} - {\bf v}_k^{\mathsf T}{\bf x}_n + t \\
s_{nk}' & \mapsto & s_{nk}' + t .
\end{eqnarray}
So which of these factor graphs is best?  Since we are interested in symmetries that affect inference, it ultimately depends on what inference algorithm is being used.  If the algorithm treats $u_{nk}$ as a separate variable with its own approximate distribution, then we should include it in the factor graph and the extra symmetries above will be genuine symmetries of the algorithm.  On the other hand, if the inference algorithm analytically marginalizes out $u_{nk}$ (as most would), then $u_{nk}$ should not appear in the factor graph and these extra symmetries will not occur.

\subsection{Difficulty Versus Ability}

The difficulty versus ability model is a generative model for how students perform on multiple-choice tests \citep{Bachrach2012}. Each question~$q$ has a latent difficulty~$d_q$, a level of discrimination~$\tau_q$, and a correct answer~$y_q$. Each student~$p$ has a latent ability~$a_p$. The advantage of student~$p$ over question~$q$ is~$t_{pq} = a_p - d_q$. A noisy version of the advantage is computed as 
\begin{equation}
t_{pq}' \sim \mathcal N(t_{pq}, \tfrac{1}{\tau_q}) .
\end{equation}
Student~$p$ then answers question~$q$ correctly if~$t_{pq}' \ge 0$. Otherwise, the students response~$r_{pq}$ is random. The factor graph for this model is shown in \figref{fig:difficulty_ability_factor_graph}.

This model has no permutation or sign-flip symmetries. Our algorithm finds the scaling symmetries
\begin{eqnarray}
t_{pq} & \mapsto & r t_{pq} \\
t_{pq}' & \mapsto & r t_{pq}' \\
\tau_{q} & \mapsto & r^{-2} \tau_q \\
a_p & \mapsto & r a_p \\
d_q & \mapsto & r d_q
\end{eqnarray}
for any~$r \in \mathbb R_+$ and the translation symmetries
\begin{eqnarray}
a_p & \mapsto & a_p + t \\
d_q & \mapsto & d_q + t
\end{eqnarray}
for any~$t \in \mathbb R$.

\subsection{Collaborative Filtering}

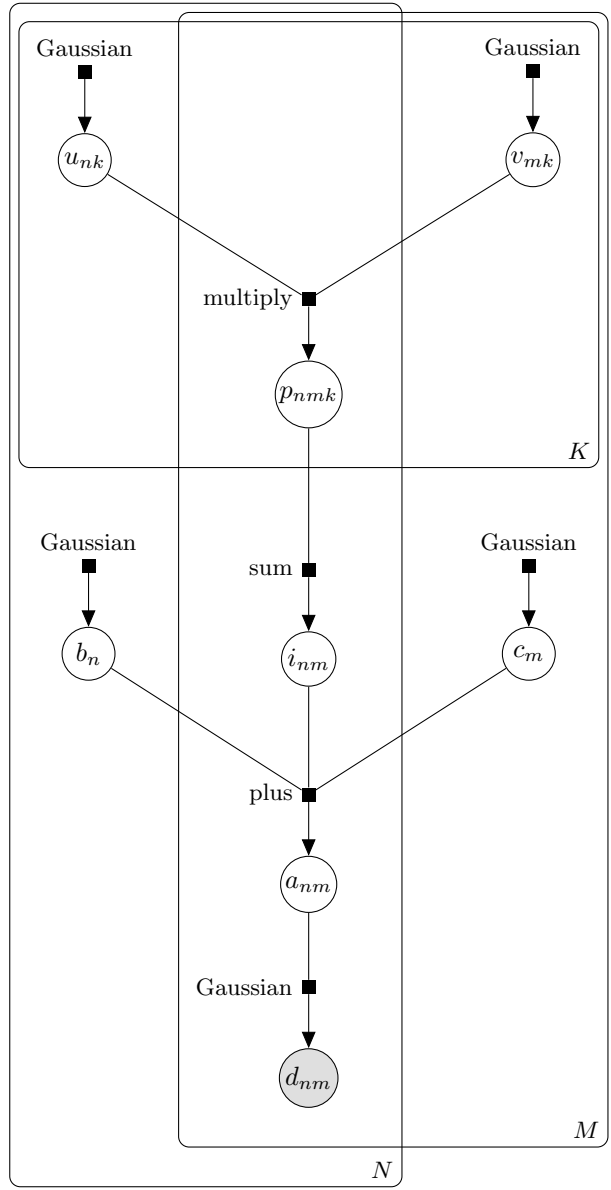
\begin{figure}[!ht]
  \begin{center}

\begin{tikzpicture}[x=1.7cm,y=1.8cm]


  \node[obs]                            (data)  {$d_{nm}$} ; %
  \node[latent, above=of data]       (affinity)  {$a_{nm}$} ; %
  \node[latent, above=1.25 of affinity]      (innerProducts)  {$i_{nm}$} ; %
  \node[latent, above=1.5 of innerProducts]       (products) {$p_{nmk}$} ; %
  \node[latent, above right=2 of products]       (items)      {$v_{mk}$} ; %
  \node[latent, above left=2 of products]       (users)      {$u_{nk}$} ; %
  \node[latent, above left=2 of affinity]   (userBiases) {$b_{n}$} ; %
  \node[latent, above right=2 of affinity]   (itemBiases) {$c_{m}$} ; %

  \factor[above=of products] {products-f} {left:multiply} {users, items} {products} ; %
  \factor[above=of innerProducts] {innerProducts-f} {left:sum} {products} {innerProducts} ; %
  \factor[above=of affinity] {affinity-f} {left:plus} {innerProducts, userBiases, itemBiases} {affinity} ; %
  \factor[above=of data] {data-f} {left:Gaussian} {affinity} {data} ; %
  \factor[above=of userBiases] {userBiases-f} {Gaussian} {} {userBiases} ; %
  \factor[above=of itemBiases] {itemBiases-f} {Gaussian} {} {itemBiases} ; %
  \factor[above=of users] {users-f} {Gaussian} {} {users} ; %
  \factor[above=of items] {items-f} {Gaussian} {} {items} ; %

  \plate {feature} { %
    (users)(users-f)(users-f-caption) %
    (items)(items-f)(items-f-caption) %
    (products)(products-f)(products-f-caption) %
  } {$K$} ;
  \plate {items} { %
    (feature.north east)
    (items)(items-f)(items-f-caption) %
    (products)(products-f)(products-f-caption) %
    (innerProducts)(innerProducts-f)(innerProducts-f-caption) %
    (affinity)(affinity-f)(affinity-f-caption) %
    (itemBiases)(itemBiases-f)(itemBiases-f-caption)
    (data)(data-f)(data-f-caption)
    (response) %
  } {$M$} ;
  \plate {users} { %
    (feature.north west)
    (items.north west)
    (items.south)
    (users)(users-f)(users-f-caption) %
    (products)(products-f)(products-f-caption) %
    (innerProducts)(innerProducts-f)(innerProducts-f-caption) %
    (affinity)(affinity-f)(affinity-f-caption) %
    (userBiases)(userBiases-f)(userBiases-f-caption)
    (data)(data-f)(data-f-caption)
    (response) %
  } {$N$} ;

\end{tikzpicture}

  \end{center}
  \caption{Factor graph for collaborative filtering.}
  \label{fig:collaborative_filtering_factor_graph}
\end{figure}

The collaborative filtering model that we consider maintains latent
features~$u_{n1},\ldots, u_{nK}$ for each of~$N$ users and latent
features~$v_{m1}, \ldots, v_{mK}$ for each of~$M$ items. There is also
an~$N$-dimensional vector of user biases~$(b_1, \ldots, b_N)$ and
an~$M$-dimensional vector of item biases~$(c_1, \ldots, c_M)$. The
affinity of user~$n$ for item~$m$ is given by
\begin{equation}
  a_{nm} = b_n + c_m + \sum_k u_{nk} v_{mk} .
\end{equation}
Noisy versions of some of the affinities, given by
\begin{equation}
  d_{nm} \sim \mathcal N(a_{nm}, \sigma^2) ,
\end{equation}
are observed. The factor graph for this model is shown in \figref{fig:collaborative_filtering_factor_graph}.

Our algorithm finds the permutation symmetries in which we permute any of the underlying features indexed by~$k$. Our algorithm finds the scaling symmetries given by
\begin{eqnarray}
u_{nk} & \mapsto & r_{nk} u_{nk} \\
v_{nk} & \mapsto & r_{mk}^{-1} v_{mk} ,
\end{eqnarray}
for all~$r_{nk}, r_{mk} \in \mathbb R_+$, such that~$r_{nk} = r_{mk}$ for all~$n, m, k$. Similarly, we find the sign-flip symmetries
\begin{eqnarray}
u_{nk} & \mapsto & s_{nk} u_{nk} \\
v_{nk} & \mapsto & s_{mk} v_{km} ,
\end{eqnarray}
for all~$s_{nk}, s_{mk} \in \{ \pm 1\}$, such that~$s_{nk} = s_{mk}$ for all~$n, m, k$.

The model's translation symmetries are a bit more complex. Note that the summation factor introduces the constraint
\begin{equation}
t_{a_{nm}}  = \sum_k t_{u_{nk}} v_{mk} \,\,\,\, \text{or} \,\,\,\, t_{a_{nm}}  = \sum_k u_{nk} t_{v_{mk}} ,
\end{equation}
which introduces constraints relating the translation constants of elements of the same array. In the first case, we apply the most restrictive form of our algorithm in which the~$t_{u_{nk}} v_{mk}$ are constrained to all be the same and the~$u_{nk} t_{v_{mk}}$ are constrained to all be the same. In this situation, we find the translation symmetries
\begin{eqnarray}
b_n & \mapsto & b_n + t \\
c_m & \mapsto & c_m - t ,
\end{eqnarray}
for all~$t \in \mathbb R$. These symmetries are a subset of the model's translation symmetries. This misses the set of translation symmetries
\begin{eqnarray}
u_{nk} & \mapsto & u_{nk} + t_{k} \\
c_m & \mapsto & c_m - \sum_k t_k v_{mk} 
\end{eqnarray}
and
\begin{eqnarray}
v_{mk} & \mapsto & v_{mk} + t_{k} \\
b_n & \mapsto & b_n - \sum_k t_k u_{nk} .
\end{eqnarray}
By ignoring the constraint imposed by the summation factor and solving the remaining system of equations, we can find a superset of the translation symmetries that includes these omitted symmetries. However, there will be some spurious solutions, and there will be remaining work to do in pruning away the solutions that are inconsistent with the omitted constraint. For any particular factor, (such as the~$n$-ary sum), we may be able to automate this pruning procedure.

Collaborative filtering is an interesting example because this model in fact contains a much broader form of rotational symmetry. For any invertible~$K \times K$ matrix~$M$, the transformation
\begin{eqnarray}
U & \mapsto & UM \\
V & \mapsto & V (M^{-1})^{\mathsf T}
\end{eqnarray}
is a symmetry of the model. The permutation, scaling, and sign-flip
symmetries that we find are special cases of this
matrix-multiplication symmetry. These matrix multiplication symmetries
are local, and we could devise a class of transformations encompassing
them (for instance, by allowing the components of the transformations
to be linear combinations of the variables in the model). We would even
be able to represent the constraints imposed by the factors as a
disjunction of linear constraints. Doing this would require
introducing a large number of auxiliary variables, and solving the
system would potentially be expensive, but it could be done. However,
the presence of such a class of symmetries will imply the presence of
various permutation, scaling, and sign-flip symmetries as in the
collaborative filtering example, all of which are much simpler to
detect.

\FloatBarrier
\section{Discussion} \label{sec:discussion}

Parameter symmetries are ubiquitous, and the presence of symmetries in a model can degrade the quality and interpretability of inference. Therefore, symmetries present an obstacle to the success of general-purpose probabilistic programming. In the context in which many non-expert machine learning practitioners seek to create novel models to describe their specific research problems, we will not be able to rely on every practitioner to be familiar with the problem of parameter symmetries or to know how to find them. As such, it will be important to have mechanisms in place for automatically detecting the presence of parameter symmetries and modifying the model so as to break the symmetries.

With this goal in mind, we introduced the concept of local parameter
symmetries and described a procedure for constructing algorithms to
automatically detect the presence of these symmetries. We illustrated
this procedure by deriving algorithms for automatically detecting
scaling, sign-flip, and translation symmetries. We discussed more
general types of symmetries, the most common of which are permutation
symmetries arising in mixture models. Our implementation of these
algorithms works for models specified in Infer.NET
\citep{InferNET2012}, but the algorithms are general and can be easily
adapted to any model specification language that arises in the future.

Graphical models can be massive, especially when we have large
quantities of data or when the models are specified with plate
notation. In these situations, we described how our algorithms can
accommodate models specified with for loops and arrays and how the
complexity of our symmetry detection algorithms is independent of the
sizes of the arrays in the model. However, the fact that our
algorithms do not depend on the sizes of the arrays in the model means
that we compromise the exactness of our algorithms in some
settings. As we seek to represent symmetries by maintaining only a
single value for each array, our algorithms have trouble finding
symmetries that require specifying the interactions between different
elements of the same array. This is primarily a problem when it comes
to representing translation symmetries in models with that use
the~$n$-ary sum factor. In such cases, we return a subset of the
translation symmetries of the model.  In total, however, our methods
are powerful enough to detect most of the symmetries that we have
found to arise in practice, as is demonstrated in our empirical study.

One area of future work will be to restore the exactness of our algorithms in the presence of arrays. Ideally, our algorithms would scale with the length of the model specification (regardless of whether or not the specification uses arrays). In such cases, when the arrays used in the models do not correspond to repetition in the model, our algorithms would first have to unroll the factor graph to the point where sufficient repetition exists for our algorithms to behave exactly.

The most important area of future work is combining the methods of this paper with automated techniques for breaking symmetries. In conjunction with specific techniques from the literature on breaking model symmetries, these results can enable the automatic detection and removal of parameter symmetries without any effort on the part of the modeler. Though much work has been done on breaking symmetry in specific models, there is work to be done in automating the procedure.

\subsection*{Acknowledgements}
We would like to thank Chris Maddison and Boris Yangel for insightful discussions.

\bibliographystyle{abbrvnat}
\bibliography{refs}

\end{document}